%% file: 0-main.tex
\documentclass[sigconf]{acmart}

\usepackage{multirow}
\usepackage{booktabs}
\usepackage{subfiles}
\usepackage{ifthen}
\usepackage{color}
\usepackage{wrapfig}
\usepackage{listings}
\usepackage{xcolor}
\usepackage{tikz}
\usepackage{listings}
\usepackage{amsmath}
\usepackage{natbib}
\usepackage[symbol]{footmisc}
\usepackage{subfig}

\newcommand{\xhdr}[1]{\vspace{3mm}\noindent{{\bf #1.}}}

\newcommand{\omt}[1]{}

\newcommand{\showComments}{false}

\newboolean{comments}
\setboolean{comments}{\showComments}
\ifthenelse{\boolean{comments}} {
	\newcommand{\sid}[1]{\textcolor{blue}{(Sid: #1)}}
	\newcommand{\reid}[1]{\textcolor{magenta}{(Reid: #1)}}
	\newcommand{\ashton}[1]{\textcolor{red}{(Ashton: #1)}}
	\newcommand{\jon}[1]{\textcolor{green}{(Jon: #1)}}
} {
	\newcommand{\sid}[1]{}
	\newcommand{\reid}[1]{}
	\newcommand{\ashton}[1]{}
	\newcommand{\jon}[1]{}
}

\def\Prf{{\rm Pr}}

\newcommand{\Prb}[1]{
\Prf\left[{#1}\right]
}

\def\ve{\varepsilon}

\ccsdesc[500]{Human-centered computing~Empirical studies in collaborative and social computing}

\begin{document}

\title{Learning Models of Individual Behavior in Chess}

\author{Reid McIlroy-Young}
\email{reidmcy@cs.toronto.edu}
\affiliation{%
  \institution{
  University of Toronto}
  \city{Toronto}
  \state{Ontario}
  \country{Canada}
}

\author{Russell Wang}
\affiliation{%
  \institution{
  University of Toronto}
  \city{Toronto}
  \state{Ontario}
   \country{Canada}
}

\author{Siddhartha Sen}
\affiliation{%
  \institution{Microsoft Research}
  \city{New York City}
  \state{New York}
  \country{USA}
}
\author{Jon Kleinberg}
\affiliation{%
  \institution{
  Cornell University}
  \city{Ithica}
  \state{New York}
  \country{USA}
}

\author{Ashton Anderson}
\affiliation{%
  \institution{
  University of Toronto}
  \city{Toronto}
  \state{Ontario}
   \country{Canada}
}
\copyrightyear{2022} 
\acmYear{2022} 
\setcopyright{acmcopyright}\acmConference[KDD '22]{Proceedings of the 28th ACM SIGKDD Conference on Knowledge Discovery and Data Mining}{August 14--18, 2022}{Washington, DC, USA}
\acmBooktitle{Proceedings of the 28th ACM SIGKDD Conference on Knowledge Discovery and Data Mining (KDD '22), August 14--18, 2022, Washington, DC, USA}
\acmPrice{15.00}
\acmDOI{10.1145/3534678.3539367}
\acmISBN{978-1-4503-9385-0/22/08}

\begin{abstract}
	\subfile{1-abstract}
\end{abstract}
\maketitle
\subfile{2-intro}
\subfile{3-related_work}

\subfile{4-data}
\subfile{5-methods}
\subfile{6-results}
\subfile{7-discussion}

\bibliography{cites}
\bibliographystyle{acm}

\clearpage

\subfile{8-supplement}

\end{document}

%% file: 1-abstract.tex
AI systems that can capture human-like behavior are becoming increasingly useful in situations where humans may want to learn from these systems, collaborate with them, or engage with them as partners for an extended duration. In order to develop human-oriented AI systems, the problem of predicting human actions---as opposed to predicting optimal actions---has received considerable attention. Existing work has focused on capturing human behavior in an aggregate sense, which potentially limits the benefit any particular individual could gain from interaction with these systems. We extend this line of work by developing highly accurate predictive models of \emph{individual} human behavior in chess. Chess is a rich domain for exploring human-AI interaction because it combines a unique set of properties: AI systems achieved superhuman performance many years ago, and yet humans still interact with them closely, both as opponents and as preparation tools, and there is an enormous corpus of recorded data on individual player games. Starting with \textit{Maia}, an open-source version of \textit{AlphaZero} trained on a population of human players, we demonstrate that we can significantly improve prediction accuracy of a particular player's moves by applying a series of fine-tuning methods. Furthermore, our personalized models can be used to perform stylometry---predicting who made a given set of moves---indicating that they capture human decision-making at an individual level. Our work demonstrates a way to bring AI systems into better alignment with the behavior of individual people, which could lead to large improvements in human-AI interaction.



%% file: 2-intro.tex
\section{Introduction}



The advent of machine learning systems that surpass human ability in various domains raises the possibility that people could learn from and interact with superhuman AI. However, such human-AI interaction is currently made difficult by the fact that algorithmic agents typically behave very differently from humans. The actions, techniques, or styles that work well for AI often do not translate to how people think. How can we construct AI models in a superhuman domain that people can actually benefit from? To bridge this gap, a natural idea is to focus on characterizing human behavior---instead of approximating optimal policies in a given domain, learning to approximate \emph{human} policies. 
Developing the ability to model human behavior in this way could provide a path toward building algorithmic learning tools that can guide people to performance improvements, or machine learning systems that humans can more easily collaborate with to achieve a shared goal.

Recent progress has been made towards this objective in the ideal model system of chess~\cite{anderson2017assessing,maiakdd}. There are a number of properties that make chess an attractive domain to pursue these questions in. First, chess AI definitively surpassed human chess-playing ability in 2005, yet millions of people still play it. There are billions of games played online each year, in which people face dozens of decision-making situations per game, and every action they take is digitally recorded. Chess has also been a leading indicator in AI and machine learning for decades. Recently, AlphaZero revolutionized algorithmic game-playing with a novel deep reinforcement learning framework. Subsequent work trained Maia, an AlphaZero-like framework, on millions of human games to predict which move a player of a given skill level will make~\cite{maiakdd}. By training several models, each on a subset of games limited to a coarse skill level, this approach captures aggregate human behavior in chess at different levels of strength. And by modeling human decision-making behavior in a domain where algorithms are already dominant, we can begin to tighten the connection between typical human behavior and superhuman AI.

Although this was an important step, the ultimate realization of modeling human behavior would be the ability to capture decision-making style at the \emph{individual} level. A model that could faithfully capture a particular person’s actions would be of clear use for automating different forms of interaction with them, such as potentially teaching them how to improve. For example, a coach that understands how \emph{you} could improve in particular would clearly be more helpful than one who only understands people at your level in general. However, in chess, just as in other domains, such as medical diagnosis and text generation, strong AI performance is often achieved by aggregating data over many people; it is far from clear whether the variation among individuals provides sufficiently distinctive signals to enable individualized models to do significantly better than these aggregate models. Indeed, our hypothesis was that this would not be true.

In this paper, we use chess as an application domain to demonstrate how to fine-tune a deep neural network to an individual person's behavior when there are thousands of examples per person, and explore the ramifications of this ability. 
In particular, we construct models of individual human decision-making in chess that significantly outperform previous aggregate population-level models~\cite{maiakdd} by a significant margin, raising move prediction accuracy for an individual player on average by over 4 percentage points. 
We achieve this by taking the open-source AlphaZero-like framework of~\cite{maiakdd} and applying recent fine-tuning and transfer learning methods to personalize move predictions to individual players. 

The resulting personalized models capture individual decision-making style. In addition to the aforementioned 4 percentage point gain---which is close to the gap in move-matching performance between unpersonalized Maia and traditional chess engines that do not even attempt to match human moves---our \emph{transfer Maia} models have much better perplexity on the target player's moves. 
Furthermore, personalized models outperform the base models across the entire spectrum of move quality---everything from good moves to serious blunders are more accurately predicted by transfer Maia. 
Most strikingly, our personalized models model individual decision-making to such an extent that we can use them to uniquely identify a player from their moves alone. In this version of the ``author attribution’’, or \emph{stylometry}, task, we are given a set of games, and the goal is to predict who played it. 
Given one side of 100 games, we can correctly identify the player who was playing 98\% of the time out of a pool of 400 players---despite \emph{not even having trained to do this task}. 
Even more, our models can perform this identification of players from their mistakes alone, meaning that they understand how each individual player differs in what they need to fix in order to improve.
We achieve similar results even when we only consider the latter parts of the game (which typically contain unseen positions), suggesting that our models capture unique aspects of a player’s style. 

In summary, we find that the ramifications of modeling an individual's decision style result in models that could power algorithmic teaching tools---ones who know your behavior so well that they can identify you from your decisions alone, and know what you specifically need to do to improve. As AI continues to rapidly progress, our work shows the value in modeling individual decision-making styles in order to align algorithms with people. 


%% file: 3-related_work.tex
\section{Related Work}

Our work applies methods that were mainly developed in the transfer learning literature, but are also closely related to imitation learning, domain adaptation, meta-learning, and multitask learning. In particular, we experiment with fine-tuning our model by freezing its bottom layers~\cite{zoph2016transfer,tajbakhsh2016convolutional,oquab2014learning}, initializing the top layers randomly versus starting from a pre-existing model~\cite{donahue2014decaf}, and varying the pre-existing model that we start with~\cite{kornblith2019better,chatfield2014return}. We derive inspiration from computer vision tasks that specialize a pre-existing model (e.g.\ Resnet-50 \cite{george2018classification}) to a specific task \cite{yosinski2014transferable, raghu2019transfusion, huh2016makes,litjens2017survey}, a method that has also been extended to many other domains, such as natural language processing \cite{zoph2016transfer,dai2015semi,yang2017neural,abs-1802-05365,howard2018universal} and speech recognition \cite{huang2016unified,kunze2017transfer,deng2013sparse,wang2015transfer}.  
Many developments in transfer learning are geared towards dealing with data scarcity by minimizing the number of samples required \cite{kunze2017transfer,rohrbach2013transfer,sun2019meta}. 
One of our contributions is to map out which techniques work best when even personalized models are relatively data-rich, a less studied setting~\cite{raffel2019exploring}. Other approaches include adding additional layers \cite{tan2018survey} or additional inputs \cite{kawahara2016multi}.

Several other machine learning tasks are closely related to our problem. Our problem could be cast as a meta-learning task \cite{finn2017model,raghu2019rapid,raghu2019transfusion,andrychowicz2016learning}, with the goal of picking the closest model to each individual. Our goal of training a model on human behavior is reminiscent of imitation learning, although a key difference is that we are starting from already-superhuman AI and aiming to design more human-oriented models, whereas imitation learning models generally aim to improve by emulating a human expert~\cite{ding2019goal,ho2016generative,torabi2019recent,schaal1999imitation}. Finally, our results on identifying players draw on some of the earliest uses of neural networks for handwriting recognition~\cite{lecun1998gradient} and performing stylometry~\cite{graves2008novel,plamondon2000online,tappert1990state,pham2014dropout,revow1996using}. The behavioral stylometry task has also been done in chess~\cite{mcilroy2021detecting}, although our previous method was trained explicitly and could only be used for player identification and works via an embedding. In contrast, here we develop personalized models that can predict each move and achieve stylometry as a byproduct of the accuracy of these models.

Our goal of capturing individual style finds its closest counterpart in the opponent modeling literature~\cite{brown1951iterative,billings1998opponent,davidson2000improved}. This line of research has focused on building a model of an opponent (e.g., in poker) in order to predict likely next actions that one could best-respond to. Our work differs in objective and approach. Instead of modeling people to exploit their weaknesses and achieve better performance, we are trying to model people so that an already-superhuman AI can better teach and interact with them. In addition, most of the opponent modeling literature employs approaches involving cardinal~\cite{lockett2007evolving} or mixture-of-experts~\cite{he2016opponent} strategies, whereas we apply a residual network deep learning framework. 

Our application domain, chess, has been used as a model system for artificial intelligence~\cite{mccarthy-chess-drosophila,newell1958chess,anderson2017assessing,haworth2009performance,paulsen2010moderately} and understanding human behavior~\cite{chase1973perception,simon1977structure,biswas-regan-icmla15,charness-psych-chess-review,gobet1996templates,regan2011intrinsic} for decades. A period of fervent work on computer chess culminated in Deep Blue defeating Garry Kasparov in 1997, but more recently the introduction of AlphaZero, a system using deep residual networks, revolutionized the state of the art~\cite{silver2018general,maiakdd,czech2019learning}. While AlphaZero is designed to approximate optimal play in chess even more perfectly than its predecessors, we adapt it to characterize human play at an individual level. Work done before AlphaZero to capture style in chess was much more limited in scope; for example one attempt using GANs considers only the first few moves and a single player's style~\cite{chidambaram2017style}. More recent papers have pursued similar goals in cooperative game-playing~\cite{carroll2019utility}, card-playing~\cite{baier2018emulating}, and chess~\cite{maiakdd}. There are also commercial products (such as \textit{Play Magnus} and \textit{Chess.coms}'s personalized bots) that attempt to mimic specific players. Their methods are not publicly disclosed, but are thought to be simple attenuated versions of existing chess engines~\cite{play_magnus}.

%% file: 4-data.tex
\section{Data and Background}
\label{sec:data}

\xhdr{Lichess} We use data from the largest open-source online chess platform, \href{https://lichess.org}{\textit{Lichess}} \cite{lichess}. With almost 100 million games played per month and almost 2 billion games played in total, Lichess provides us with a large number of diverse chess players, some of whom have played tens of thousands of games each. Games are played at a variety of set durations ranging from long games, where each player has an hour or more, to extremely quick games, where each player has only 15 seconds for the entire game. For our work we ignore the fastest games, as players tend to make many more mistakes, sometimes intentionally, in order to not run out of time. We analyze games from the Blitz category, where players have between 3 to 8 minutes per game. Each player has a rating~\cite{glickman1995glicko} that represents their skill level, and is derived from their results against other players on the platform. The rating system is calibrated such that a player who outrates their opponent by 200 points is expected to win 75\% of the time. As most players are rated between 1100 and 2000 on Lichess, we restrict our attention to these players only. Lichess has a robust community with impressive capabilities to detect bots and cheaters who use chess engine assistance. Players are also encouraged to maintain a single continuous account. As a result, we are able to train models on hundreds of human players with over 20,000 games played each.

\xhdr{Dataset construction} To assemble a specific set of players to train personalized models on, we first collected a dataset containing all rated games played on Lichess between January 2013 and December 2020. We then defined a set of criteria to select players: at least 20,000 games played, mean rating between 1000 and 2000 (for consistency with~\cite{maiakdd}), low variance in rating, at least one game played in December 2020, and account older than one year. See our supplement for the full details. We then grouped players by the number of games they have played, and randomly assigned players from these groups into exploration (10\%), evaluation (80\%), and holdout (10\%) sets. The exploration set is used to configure the training parameters and architecture of our fine-tuning methodology. This methodology is then applied to the evaluation set to train and test personalized models for those players. 
The median rating of players in the exploration and evaluation sets is 1750 and 1739 respectively. The holdout set was never used for this work and is reserved for future analysis. Table~\ref{train_test_counts} shows the composition of the player sets.

For each player, we randomly split their games into four datasets: \textit{training}, \textit{validation}, \textit{testing}, and \textit{future}. The first three are standard splits of all games played before December 2020, randomly sampled to yield 80\%, 10\%, and 10\% splits respectively. The \textit{future} dataset contains all games played in December 2020, which occur later in time compared to all other datasets. By splitting at the game level, we allow for positions to be shared between the training and test sets. The opening board is shared by all games, but by ply 20 (a ``ply'' is a move made by one player) the percentage of positions shared between sets drops below 10\%. Figure \ref{ply_in_out_sample} shows the ratio of new positions by ply.


 \begin{table}[t]
    \centering
    \caption{The number of players in each of our player sets, grouped by the minimum number of games played. 
   \\ \textit{*} 40 random players were selected from these sets. \textdagger Unused}
     \subfile{tables/train_test}

    \label{train_test_counts}
 \end{table}

\xhdr{Leela and Maia}\label{value_func} Our work builds on two chess engines projects. The first, \href{https://lczero.org}{\textit{Leela Chess Zero}}, is an open source implementation of the deep reinforcement learning system \textit{AlphaZero Chess} \cite{silver2018general}, and provides the code infrastructure that we leverage to build our models. The second, \textit{Maia Chess}, is a supervised learning adaptation of Leela that attempts to predict the next move the average human player at a particular rating level will make in a given chess position~\cite{maiakdd}. Maia can predict human moves much more accurately than the previous state of the art, attenuated versions of strong chess engines such as Stockfish and Leela. There are 9 versions of Maia, one trained on each rating level from 1100 to 1900. 

%% file: tables/train_test.tex
\begin{tabular}{rll}
\toprule
 \# Games &  Exploration &  Evaluation \\
\midrule
      1,000 &             3,852\textdagger &         30,821* \\
      5,000 &              662\textdagger &          5,298* \\
     10,000 &              233* &          1,866* \\
     20,000 &               36 &           295 \\
     30,000 &               10 &            86 \\
     40,000 &               10 &            19 \\
\bottomrule
\end{tabular}

%% file: 5-methods.tex
\section{Methodology}\label{sec:methods}


Our high-level approach is to take existing Maia models, which are designed to predict human moves at a particular skill level, and specialize them to predict the moves of an individual player. 
This type of transfer learning can be carried out in myriad ways; we organize our investigation through a logical sequence of design decisions that we explore in turn. The end result is a transfer learning methodology for creating a personalized model for any player, given a sufficient number of the player's games. 

To train our models, we used a heavily modified version of Maia Chess. Since these are Leela-based models, they use a series of residual connections~\cite{he2016deep} between convolutional layers with ReLU activations and squeeze layers~\cite{hu2018squeeze}, but without pooling layers.
The input representation is a 112-channel $8 \times 8$ board image representation. 
The output is the predicted move (policy head) which is represented by a 1858-dimensional vector. 

\subsection{Model Parameter Selection}\label{sec:params}

\xhdr{Training parameters} To understand the impact of various hyperparameter choices and design decisions, we conducted exploration in two main phases. We first performed an initial, broad set of analyses on a set of 10 exploration players with over 40,000 games each---we refer to these players as the \textit{initial set}. Then, we performed our deeper methodological experiments on the \textit{full set} of 96 exploration players: 10 with over 40,000 games each, 10 with over 30,000 games each, 36 with over 20,000 games each, and a random sample of 40 players with over 10,000 games each. We evaluated both accuracy and cross-entropy loss on the initial set using the players' validation games (and checked training dataset loss at frequent intervals). We then used the players' testing datasets for the final evaluation, which we provide here. As in previous work~\cite{maiakdd}, we primarily consider move-matching accuracy after the tenth ply (a ``ply'' is a move made by one player). As shown in Section~\ref{sec:results}, early moves (e.g.\ before the tenth ply) are significantly more predictable because players repeatedly encounter the same opening positions. Thus, our methodology is geared towards finding a transfer learning architecture that can capture human style in unseen positions, as opposed to memorizing or characterizing play in known positions. The optimizer used for all training was the Tensorflow Keras stochastic gradient descent and momentum optimizer. See the Supplement for full training code.

\xhdr{Depth of gradient flow} The two most important strategies we explored, in line with previous work, were freezing the top layers of our deep residual network, and either initializing weights randomly or with a preexisting model. We tested freezing our network at every reasonable stopping point, and generally found that the deeper the gradients flowed the better performance we achieved (see full results in the Supplement). Our initial experiments also generally showed that initializing our model with Maia's weights also tended to give a boost at deeper stopping points.

\xhdr{Initial model choice} Previous work found that the Maia models best predict players near the rating level they were trained on (e.g.\ Maia 1500, the version trained on 1500-rated players, best predicts players that are rated 1500 and 1600)~\cite{maiakdd}. Because of this fact, and that indivual move decisions are influenced by rating~\cite{regan2011intrinsic}, we expected the choice of which Maia we start with to also show this pattern---for example, that developing a personalized model for a 1500-rated player would benefit from starting with Maia 1500 as opposed to Maia 1900. Surprisingly, however, our experiments showed that our move prediction accuracy for a particular player doesn't vary much with the choice of Maia we start with, suggesting that the fine-tuning process dominates this choice. See section \ref{sec:sub_heatmap} of the supplement for the the effects on our 10 exploration players. We also tried randomly initializing weights, which had a significant negative effect, scaling with the depth of the gradient stopping point. Introducing small amounts of Gaussian noise to the weights before training also had no effect or negative effects, depending on the amount of noise. Therefore, for the rest of the paper we start from Maia 1900 for simplicity and ease of comparison, the model trained on the highest-rated population of players.

\xhdr{Number of steps} In the initial exploration, we ran the models for a large number of steps (150,000) and observed the validation loss curve. This suggested that most improvement occurred in the first 12,000 steps (3 million board-move pairs). For our final models, we used 30,000 steps with drops in learning rate at 15,000, 20,000 and 25,000, each by a factor of 10. In the final exploration we included a longer training configuration, which showed the same effect.

\xhdr{Sampling function}
As noted earlier, during our initial exploration we observed that earlier moves are more predictable, since players repeatedly encounter some opening positions. To emphasize the middlegame and endgame, we switched from sampling positions with uniform probability across the game, to sampling moves using a scaled $\beta(2,6)$ distribution over plies (dividing by 150 to normalize them to $[0,1]$). Although this change showed a minor uplift in validation accuracy for the initial set, it showed no effect during our  exploration over the full set.

\xhdr{Final tuning} Our initial analyses narrowed our search for a suitable model architecture; as a result of these explorations we decided to initialize from base Maia models instead of random initialization, we realized that always starting with Maia 1900 is best, we determined that freezing layers only hurts performance, and we optimized the number of steps and sampling function. In our deeper exploration over the full set of 96 exploration players, we optimized the learning rate. We found that for all data regimes from 10,000 to 40,000 games, a learning rate of 0.0001 is optimal. Table~\ref{explore_accuracy} in the supplement show the complete results. Adding this learning rate to the design decisions above results in our final architecture, which we will refer to as ``Transfer Maia'' for the rest of the paper. 

%% file: 6-results.tex
\section{Results}\label{sec:results}

We apply our transfer learning methodology to specialize the baseline Maia model (Maia 1900) to the 400 different players in our evaluation player set. We use 80\% of each player's games to train their personalized models, and evaluate them on a 10\% test dataset and the future dataset (the 10\% holdout dataset remains untouched). Here, we show the results for the original task of predicting the target player's moves, as well as a different stylometry task of identifying the player given only a subset of their moves. Since all of our results on the test and future datasets are qualitatively identical, we report results for the test datasets only.

\subsection{Move Prediction Accuracy}\label{sec:accuracy}
\begin{figure}[t]
	\centering
\includegraphics[width=.49\textwidth]{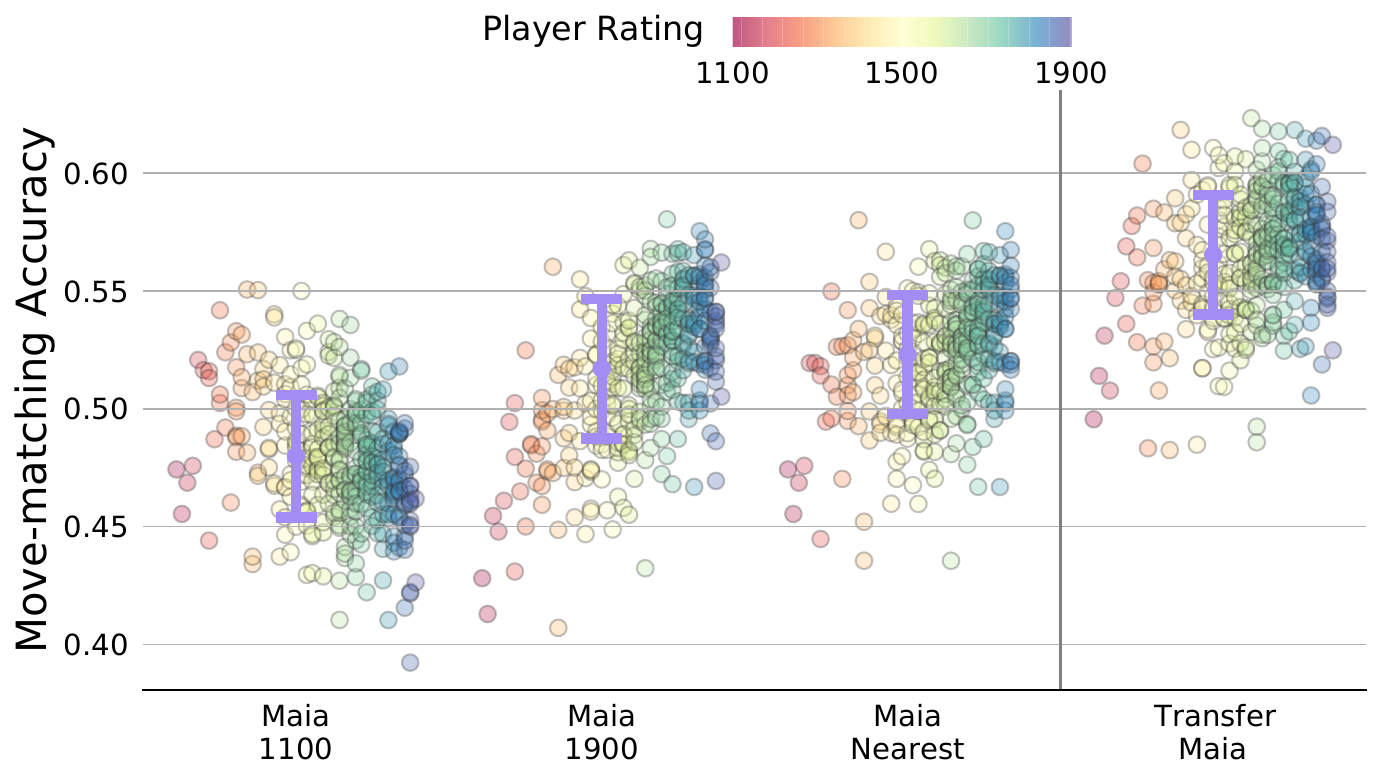}
\caption{Move-matching accuracy of aggregate Maia models (left) versus our personalized models (\emph{Transfer Maia}, right) on target players' test set games. Target players are color-coded by rating level from red (lower) to violet (higher). Maia Nearest is the Maia model whose training level is closest to the target player's rating. Error bars show one standard deviation (standard errors are too small to be visible).}
\label{main_val_accs}
\end{figure}

We evaluate how accurately our personalized models predict the moves of their target players on their testing datasets. Since the early part of the game is often repeated, we ignore the first 10 plies of each game. 
For comparison, we also evaluate the Maia models on the testing dataset; as we observed, these models represent the state-of-the-art in human move prediction, but they can only predict at coarse skill levels (chess ratings). Thus we expect the Maia model whose training level is closest to a target player's rating, which we call the ``nearest'' Maia, to have the highest baseline accuracy. Figure~\ref{main_val_accs} shows the results, where the target players are color-coded by their rating and ordered from left (lower rating) to right (higher rating). As expected, Maia's accuracy increases as its training level approaches the target player's rating, which for example explains the downward tilt of Maia 1100 and the upward tilt of Maia 1900. This replicates the main result of McIlroy-Young et al.~\cite{maiakdd}. The ``Nearest Maia'' column combines the top performing Maia models per player. In contrast, our personalized models outperform all Maia models by a sizable margin, achieving 4-5\% higher accuracy on average than the top performing Maia model, per player. That we can achieve significantly higher move prediction accuracy through personalization is a result that is neither implied by nor expected based on the performance of Maia, which has enough difficulty predicting moves at coarse rating levels.

\begin{figure}[t]
	\centering
\includegraphics[width=0.49\textwidth]{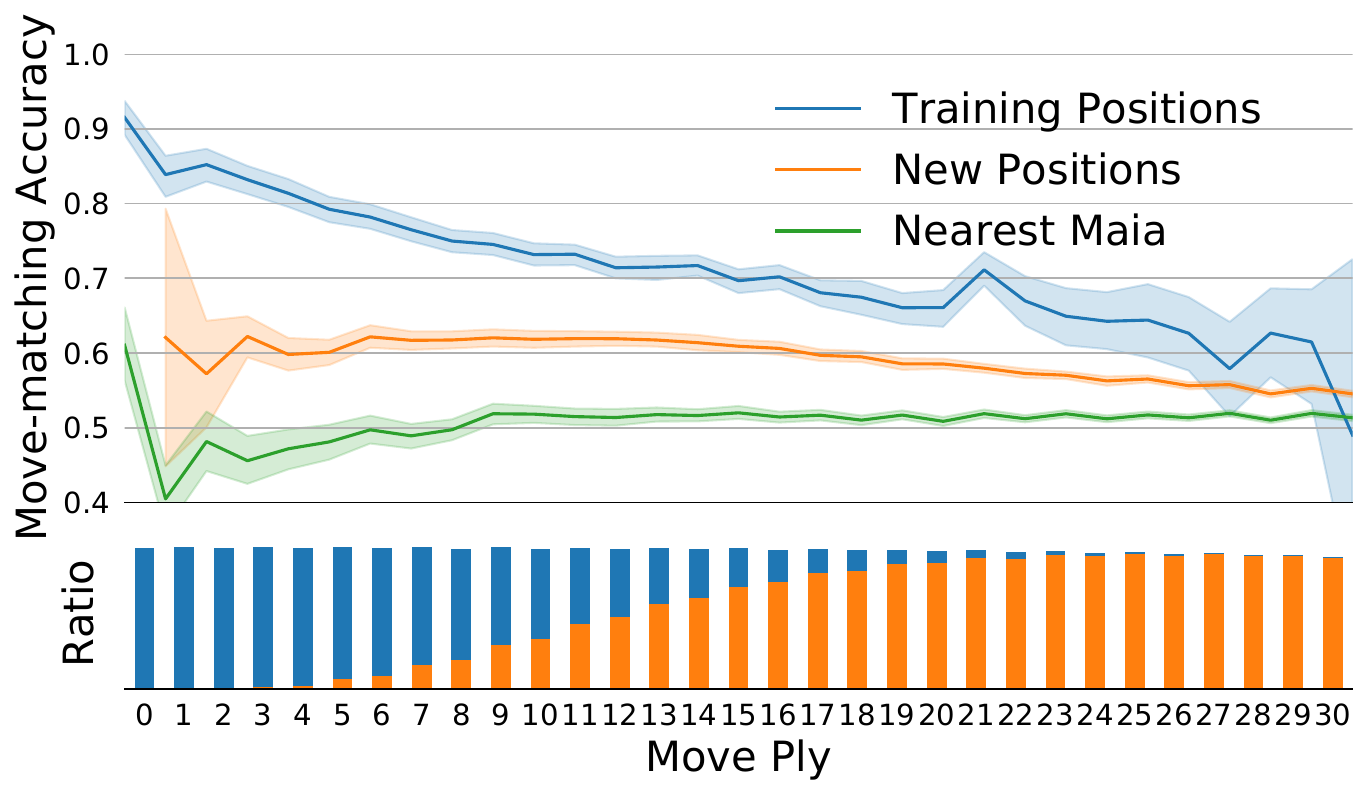}
\caption{Move-matching accuracy as a function of move ply, comparing positions that were encountered in the training data versus unseen positions. \textit{Nearest Maia} was run on all positions as a baseline. Confidence intervals are shown and calculated via bootstrap sampling. The bottom panel shows the ratio of training positions to new positions for each ply.}
\label{ply_in_out_sample}
\end{figure}

\xhdr{Game phases} A plausible explanation for the higher accuracy of our models is that they are memorizing formulaic patterns in the opening play, or other easily predictable aspects of the target player's style. To investigate this, we perform a finer analysis of model accuracy along different dimensions. Figure~\ref{ply_in_out_sample} shows how model accuracy varies as the move number in the game increases. While both the personalized models and Maia models benefit from the higher predictability of the opening play, the benefit quickly diminishes as more moves are played. Positions seen during training are predicted with higher accuracy than unseen positions. The bottom panel shows the ratio of number of training positions to new positions. Plies 0--4 (including the opening board) were seen in the training dataset 100\% of the time, but by ply 30 only 0.02\% positions were seen in the training dataset\footnote{See figure \ref{per_ply} in the supplement for an extended view}. Despite these dynamics, personalized models achieve consistently superior accuracy by a significant margin throughout the entire game.

\begin{figure}[t]
	\centering
\includegraphics[width=0.49\textwidth]{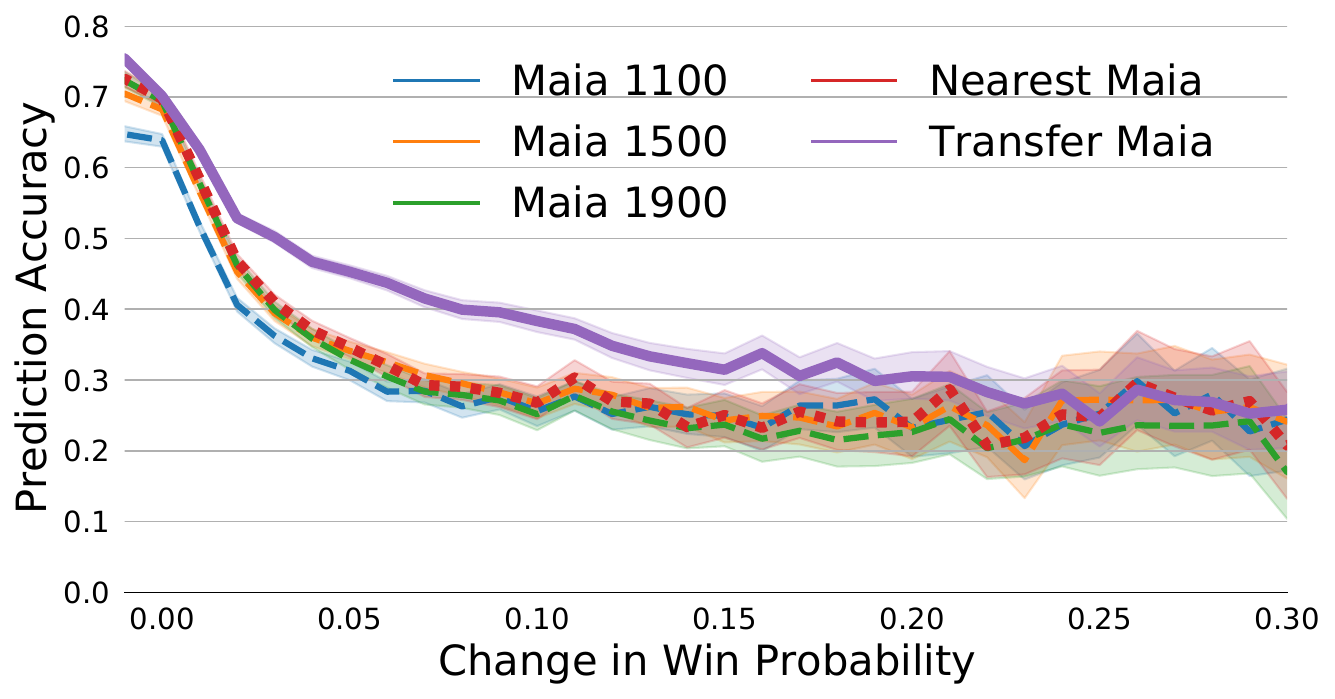}
\caption{Move-matching accuracy of our personalized models versus Maia models as we vary the quality of the move played. Our personalized model, Transfer Maia, outperforms unpersonalized baseline Maias across the entire spectrum of move quality. Confidence intervals are shown and calculated via bootstrap sampling.}
\label{winrate_loss_v_acc}
\end{figure}

\begin{figure}[t]
	\centering
\includegraphics[width=0.49\textwidth]{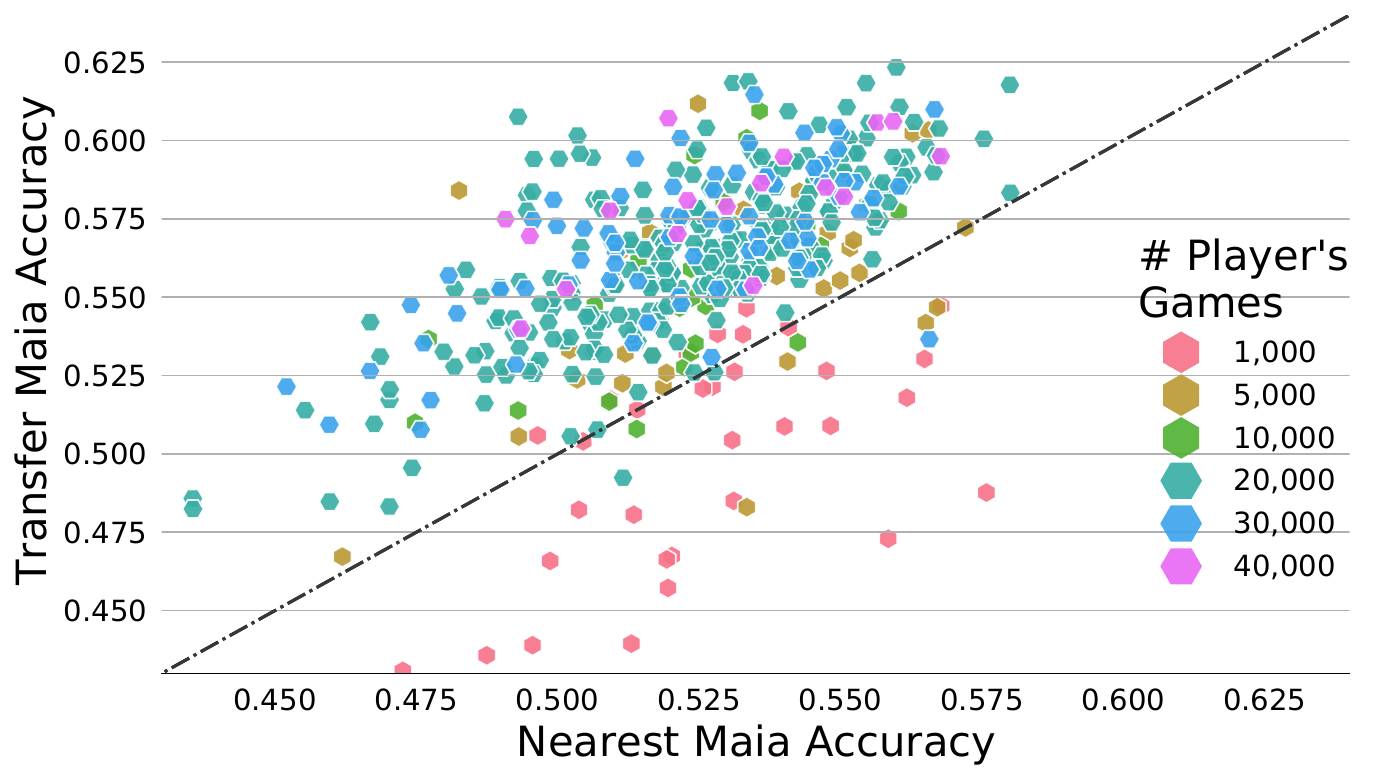}
\caption{Personalized move-matching accuracy as a function of best baseline Maia move-matching accuracy, grouped by training set size. The $y=x$ line is shown; points above it are players for whom the transfer model outperforms Maia.}  
\label{uplift_low_count}
\end{figure}

\xhdr{Move quality} One of our driving motivations behind developing personalized models of human behavior is to inform the future design of algorithmic learning tools. As such, characterizing the errors that people make is of primary interest. Figure~\ref{winrate_loss_v_acc} shows how prediction accuracy varies with the quality or ``goodness'' of the move being predicted. The quality of a move is measured by the change in estimated win probability before and after the move, where win probability is calculated via an empirical procedure based on evaluation of the position (who is ahead in material, strategic considerations, etc.) following the method of~\cite{maiakdd}. The change is always non-positive because it is measured against optimal play---the optimal move changes the win probability relative to optimal play by 0, and every worse move lowers the win probability. As the figure shows, both the personalized models and Maia models can predict better moves with higher accuracy than worse moves, but the personalized models are consistently more accurate across all move qualities by a significant margin. 
\xhdr{Training data size} Although this work is mainly concerned with demonstrating the possibility of transfer learning to capture individual decision-making style, another advantage of chess as a domain is that we can measure how training data size affects the performance of predicting individual decisions. In Figure~\ref{uplift_low_count}, we show how personalized move matching accuracy varies as a function of base Maia performance, grouped by how many games each player has played. There are two main insights to draw from this analysis. First, there is a strong correlation between the performance of these models, indicating that some players are more predictable than others. Second, personalization performance increases with training data size. Our architecture produces significant uplift over base Maia for players who have played at least 5,000 games. However, it is ineffective on players who have only played 1,000 games, many of whom base Maia performs better on (see Table~\ref{tab:datasize} for aggregate move-matching accuracy by training set size). We speculate that the lower accuracy is due to over-fitting on an inadequate training set. 

Developing effective personalization methods for players with less data, perhaps by pooling data across several players, is a fruitful direction for future research.

\begin{table}[t]
    \centering
     \caption{Move-matching accuracy for players with differing numbers of games. } 
    \begin{tabular}{rrr}
    \toprule
    \# Player's Games &  Nearest Maia   &  Transfer Maia  \\
 
    \midrule
    1,000               &         0.527 &               0.497 \\
    5,000               &         0.532 &               0.550 \\
    10,000              &         0.528 &               0.556 \\
    20,000              &         0.524 &               0.564 \\
    30,000              &         0.520 &               0.567 \\
    40,000              &         0.528 &               0.580 \\
    \bottomrule
    \end{tabular}

    \label{tab:datasize}
\end{table}

\xhdr{Skill levels} A distinguishing property of the personalized models compared to Maia is that their accuracy is largely independent of the target player's rating in the range we examine (1100--1900 Elo). The $R^2$ of an ordinary least squares regression of accuracy against player rating is 0.089, which is minor. This property may not hold across all rating levels, however: our preliminary attempts to create personalized models for Grandmasters did not yield the same accuracy gains as above. We conjecture that at the highest levels of play, it is more difficult for human-oriented models such as Maia and ours to achieve higher accuracy than chess engines like Stockfish or Leela, because the strongest players tend to play near-optimal moves which traditional chess engines are designed to predict. 


\subsection{Stylometry}
\label{sec:stylometry}

Having created personalized models that achieve higher accuracy than non-personalized baselines, a natural question to ask is how unique these models are. To what extent are we really characterizing the idiosyncrasies and signatures of individual players? Concretely, can the personalized model of one player be used to predict the moves of another player? We examined this by running each transfer model on 100 games from each other player. We can express this task for $n$ players $A_1, A_2, \ldots, A_{n}$ and we have a transfer model $M_i$ for each player $A_i$.
A model $M_i$ applied to a player $A_j$ that the model wasn't trained on ($j \neq i$)
will generally predict each move with a lower probability than it does when applied to the player $A_i$ that it was trained on.

We can think of each prediction made by any model as a $0$-$1$ random variable (with $1$ corresponding to a correct prediction and $0$ corresponding to an incorrect one); then we expect the number of correct predictions to be the highest with the correct model after a sufficient number of games. We found that 94.5\% of the time, across 100 games the correct model had the most correct predictions using 10+ ply like in the previous sections.

\begin{table}[t]
    \centering
    \begin{tabular}{lllrrr}
    \toprule
     \# games &   all moves &  10+ ply &  30+ ply  \\
    \midrule
    10 & 0.86 &  0.47 & 0.11 \\
    30 &  0.94 &  0.81&           0.26 \\
    100 & 0.98    & 0.95& 0.55  \\
    \bottomrule
    \end{tabular}
    \caption{Stylometry top-1 accuracy for various numbers of games used and different ply thresholds.}
    \label{tab:stylotable}
\end{table}

This result has profound implications, because it means that we can uniquely identify a player by simply inspecting a small sample of their games. The method is simple: given a player's games, test each personalized model on those games and output the one with the highest move prediction accuracy. Since we know the personalized models perform poorly on games outside of the player set, we can also identify when a game has not been played by any of the players by using a simple accuracy threshold. Our personalized models indeed capture individual style. 

To examine these results more closely, we show how stylometry performance among the 400 evaluation players varies with ply cutoff and number of games considered (see Table~\ref{tab:stylotable}). The later the ply cutoff, the fewer opening positions are considered, and thus the harder the prediction task is. On the other hand, accuracy increases as we consider more games. Stylometry accuracy---recovering the target player's identity from the set of 400 players with a single guess---ranges from 98.3\%, using all moves from 100 games, to 11\%, using only moves after ply 30 from 10 games. In all cases, our personalized models substantially outperform random guessing (0.25\%). As a baseline for stylometry, we train a Naive Bayes classifier (one for each game length) on the vector of centipawn losses incurred by each player on their training set games, and use this classifier to identify the most likely player given a game (and its associated centipawn loss vector). Though a reasonable style marker, this baseline achieves very poor accuracy compared to our personalized models, peaking at 1.4\%, which is only slightly better than random guessing. Other style markers may achieve better results, but finding a good style marker is difficult in general~\cite{gatys2016image}, or alternatively an embedding vector can be used~\cite{mcilroy2021detecting}. In effect, our personalized models serve as a proxy for style markers: by abstracting the differences between players through the move prediction task, we can perform stylometry without the need for explicit style markers. 

\xhdr{Blunder style} To investigate whether our models capture individual ``blunder style'', we performed a new stylometry task where we only considered ``blunders''---the roughly 35\% of all moves that decreased the player's win probability by more than 10\% (which corresponds to decreasing the Stockfish evaluation from 0.0 to $-1.7$ pawns). The result was a slight \emph{increase} in accuracy from 0.983 to 0.989, which suggests that mistakes are the most discriminative moves when it comes to identifying players. Thus, a player's personalized model makes blunders like that \emph{individual player}.

\xhdr{Stylometry bounds} 
Our stylometry results are arising from the fact that a personalized model is generally 4-5\% more accurate on the player it's trained on than on other players.
If we want to understand how a 4-5\% improvement in model performance (across the entire game) can lead to a high discriminatory power over 100 games, we can calculate some bounds on the number of games required by picking $\delta$ as an accuracy parameter and then asking: ``If we have the moves of a player $A_i$, how many moves $m$
do we need before there is a probability of at least $1 - \delta$
that model $M_i$ makes the most correct predictions?''

The computational results reported above provide quantitative answers to this question in practice; but a simplified theoretical model can provide insight into the number of moves required.
To formulate this theoretical model, let's assume that accuracies are independent and identically distributed across moves, with a model performing correctly with probability $p$ on players the model wasn't trained on, and probability $p (1+ \ve)$ on the player it was trained on.

We will also assume (consistent with the computational results) that $0.5 \leq p < 1$, and that $\ve$ is small (though we will only need it to be at most $1$ for the bounds here).  

In the appendix, we use \textit{Chernoff bounds} \cite{chernoff1952measure} for the sum of independent 0-1 random variables to show that if 
\begin{equation}
m  >  \frac{24}{\ve^2}\left[\ln n + \ln\left(\frac{1}{\delta}\right)\right]
\label{eq:m-bound}
\end{equation}
then the following two properties jointly hold with probability at least $1 - \delta$: (i) the model $M_i$ trained on player $A_i$ makes more than $pm (1 + \ve/2)$ correct predictions; and (ii) each model $M_j$ for $j \neq i$ (representing the models for each other player) makes less than $pm (1 + \ve/2)$ correct predictions.

Provided that properties (i) and (ii) both hold, the model $M_i$ will make the most correct predictions, and hence selecting the model with highest accuracy will succeed in correctly identifying player $A_i$.
When $m$ exceeds the bound specified by Equation (\ref{eq:m-bound}), this success will occur with probability at least $1 - \delta$.


\omt{

}

This allows us to calculate the number of moves required for stylometry to succeed with a certain probability. Suppose there are $n = 400$ players and the personalized models achieve $p = 0.5$ on players they weren't trained on and
$p (1 + \ve) = 0.55$ on the player they were trained on. (These numbers are representative of the real distributions; note that the personalized models perform worse than Maia when applied to other players.)
If we want at least a $0.9$ probability that the model making the most
correct predictions is the right one, then it would be sufficient to
have at least $2400(\ln(400) + \ln(100)) < 20000$ moves.

\subsection{Towards Personalized Learning Tools}
\label{sec:dist}
\begin{figure}[t]
	\centering
\includegraphics[width=0.49\textwidth]{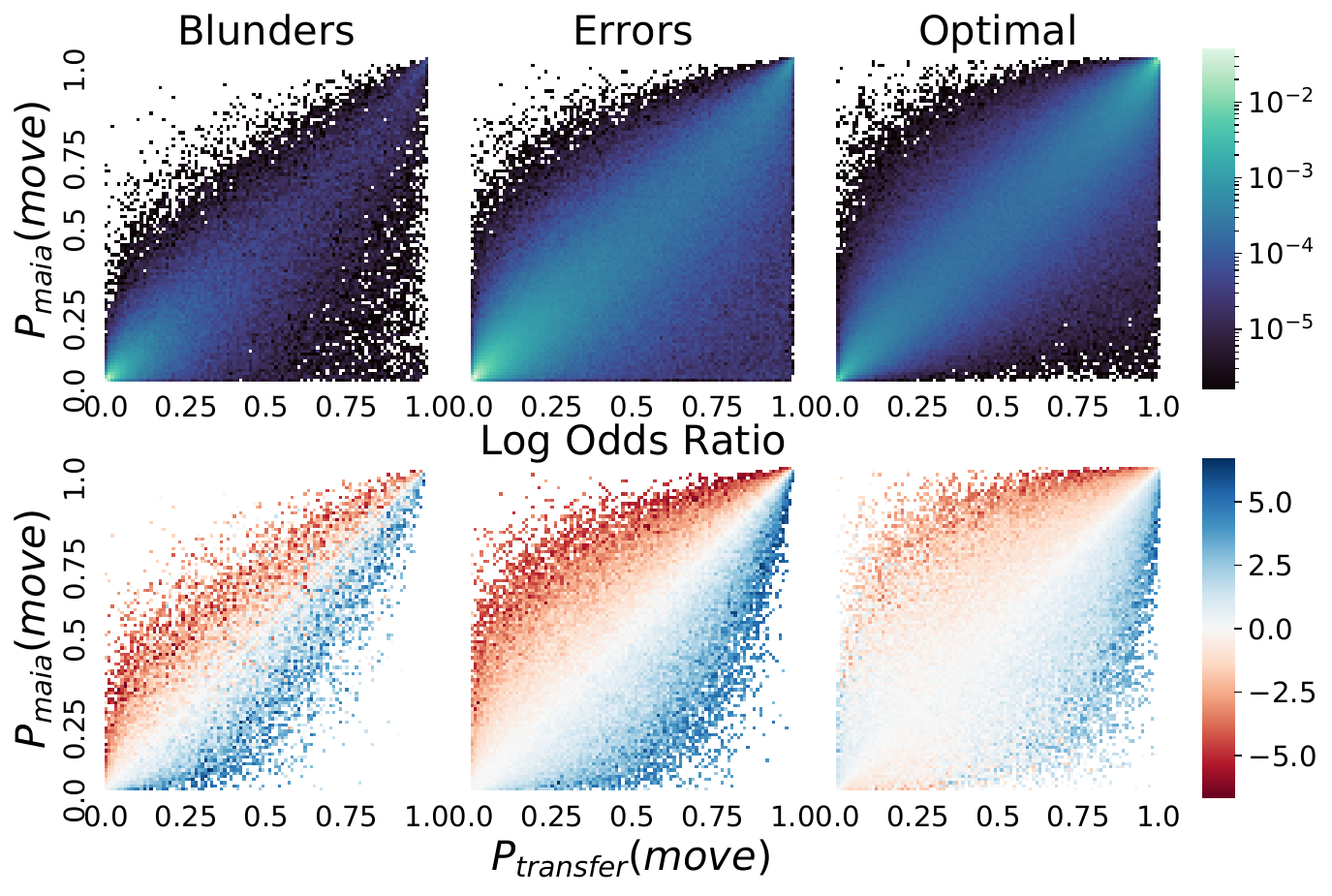}
\caption{(Top) Probability assigned to moves played by the player's personalized model ($x$) and base Maia 1900 ($y$), for large mistakes (left), small errors (middle), and optimal moves (right). (Bottom) Log odds ratio of $p(x,y)$ and $p(y,x)$. Moves above the $y=x$ diagonal are those that Maia 1900 was more certain about, whereas moves below are those the player's personalized model assigned higher probability. \textit{Optimal} moves are Stockfish's top move, \textit{Errors} are moves that aren't Stockfish's top move but only reduce the expected win-rate by at most $10\%$, and \textit{Blunders} reduce the expected win-rate by $\geq10\%$.}
\label{comparison}

\end{figure}

We have built models of human play that capture individual style. How might these types of models be useful? Here, we show that they outperform the baseline population-level Maia models on all types of decisions, including major mistakes. This implies that our technique of personalizing Maia to individual players could surface the idiosyncratic mistakes that players make. This would likely be a very useful class of moves for players to train on, since by construction they are situations the player found difficult (since they made a mistake), they are predictably difficult for this specific player (according to the output of their personalized model), and they are relatively fixable (since other players are less likely to err).

For this analysis, we consider the moves made by a player $A$, and compare the probabilities assigned to these moves by base Maia 1900 and by $A$'s personalized model. We observe that the transfer models have a mean perplexity of 1.95 while Maia 1900 has a perplexity of 2.15 (ignoring opening positions, which have a much larger perplexity difference of 0.92 and 2.40, respectively). In Figure~\ref{comparison}, we visualize how the model predictions compare for major blunders (left), minor errors (middle), and optimal moves (right). The top row shows the raw joint distribution $p(x,y)$ (where $x$ is the probability assigned by the personalized model and $y$ corresponds to base Maia), and the bottom row shows the log odds ratio comparing $p(x,y)$ to its symmetric counterpart $p(y,x)$. The virtually always-positive log odds ratio below the $y=x$ diagonal indicates that our personalized models clearly outperform base Maia on all types of decisions, from optimal moves to very large mistakes. Additionally, the larger the difference between $x$ and $y$, the more likely it is that the personalized model is the one that correctly assigns more probability to the move played. Among other learning methods, our models could be used to identify blunders for which $x>y$, as described above. 





%% file: 7-discussion.tex
\section{Discussion}

This work takes the next step towards learning {\em human} policies by building on the coarse skill-level models of prior work to create personalized models for individual players. Our models can predict a player's moves more accurately than any prior baseline, regardless of when the move occurs in a game. Furthermore, the gains we see from personalizing to individual players are significant: 4--5 percentage points higher accuracy on average than the top performing Maia model per player.

Finally, our models capture enough individual behavior to allow us to perform near-perfect stylometry. Given a sample of games, we can uniquely identify who played them from among a set of 400 players. An interesting challenge is to scale up our methodology along different dimensions: e.g., more players, more skill levels, and more time periods. For example, as we mentioned, creating accurate personalized models for the strongest players (Grandmasters) remains an elusive goal.

Machine learning systems that capture human behavior in a personalized way will open the door to algorithmic learning tools and training aids. Using chess as a model system, we have shown that predicting granular behavior at an individual level is possible. We hope this inspires others to advance human learning and facilitate collaboration with AI systems in a variety of domains.

\subsection{Ethical implications} 

Chess has traditionally been an open, public ecosystem, with both online and offline games documented, uploaded, and made easily available to the public. Our work relies entirely on this public data, and we have striven to maintain this culture of transparency by making our code and model architectures available to anyone wishing to reproduce our results. Although all the data we use is public, we nevertheless anonymize the usernames and other identifying information of all players for whom we develop personalized models.

While we believe that chess as a domain is relatively less risky than many others in which AI is applied, we believe it is valuable to highlight and consider three categories of ethical implications arising from this work.

First, powerful stylometry raises privacy concerns, because it can potentially be used to deanonymize individuals who are trying to keep their identity from being discovered. This type of deanonymization is particularly worrisome given its history of being disproportionately directed towards individuals from marginalized groups. The current work can help in exploring the privacy implications inherent in stylometry in a relatively benign arena, and to minimize the risks from this exploration. 

Second, in a more chess-specific direction, the presence of superhuman chess engines has led to the occurrence of cheating during online game play. Our personalized models do not increase the overall effectiveness of cheating in pure move quality, because predicting an opponent's next move is not as effective as knowing the optimal move to play; but they may be effective for cheating in a different sense, since they could be used to circumvent cheating detection systems. The incentive to cheat using our personalized models are not significantly greater than the baseline Maias developed in previous work. However, our personalized models could be used to target practice training with a specific player in mind, so as to increase the likelihood of defeating that player in a game.  This latter point---the potential to build specialized models of individuals---raises additional ethical questions that we believe this work can help focus and sharpen, and which we will continue exploring through our research.

Third, the development of AI applications that can imitate human behavior not just in a generalized sense but down to the level of individuals raises qualitatively new types of ethical questions, by changing the ways in which some of our basic norms and expectations around human interaction operate.  In related work, we explore the implications of such {\em mimetic models}, in which machine learning is used to design an agent that aims to imitate the behavior of a specific individual \cite{mcilroy2022mimetic}.
Such models raise a number of new possibilities that are fraught with ethical considerations: the ability to ``practice'' interactions multiple times with a model of an individual before ever meeting them; the ability to replace an individual's work with the work of a model based on them; the ability to create a model of oneself as a ``force multiplier''; and other scenarios make clear that these types of techniques raise questions that will need considerable further investigation and analysis to fully understand.

%% file: 8-supplement.tex
\section{Supplement}\label{sec:supplement}

\subsection{Code Release}

The code for training and evaluating the models is available at \texttt{\href{https://github.com/CSSLab/maia-Individual}{github.com/CSSLab/maia-individual}}, along with example games of models playing a human.

\subsection{Dataset Creation}

\subsubsection{Player Inclusion Criteria}

We constructed our player selection with the following logic: we required each player to have played over 1000 games in blitz, with a mean rating between 1000 and 2000 and a per-game rating variance under 75, have played at least one game before 2020 and one game after the first of December 2020, and we excluded titled players. Titles are given for having an official FIDE title or being a bot. We also excluded players with high or low win rates and high or low numbers of games played as white. There are also simple filters to exclude color cheating (playing one color disproportionately often, usually White) and other types of manipulation.

\begin{figure}[t]
 \includegraphics[width=.45\textwidth]{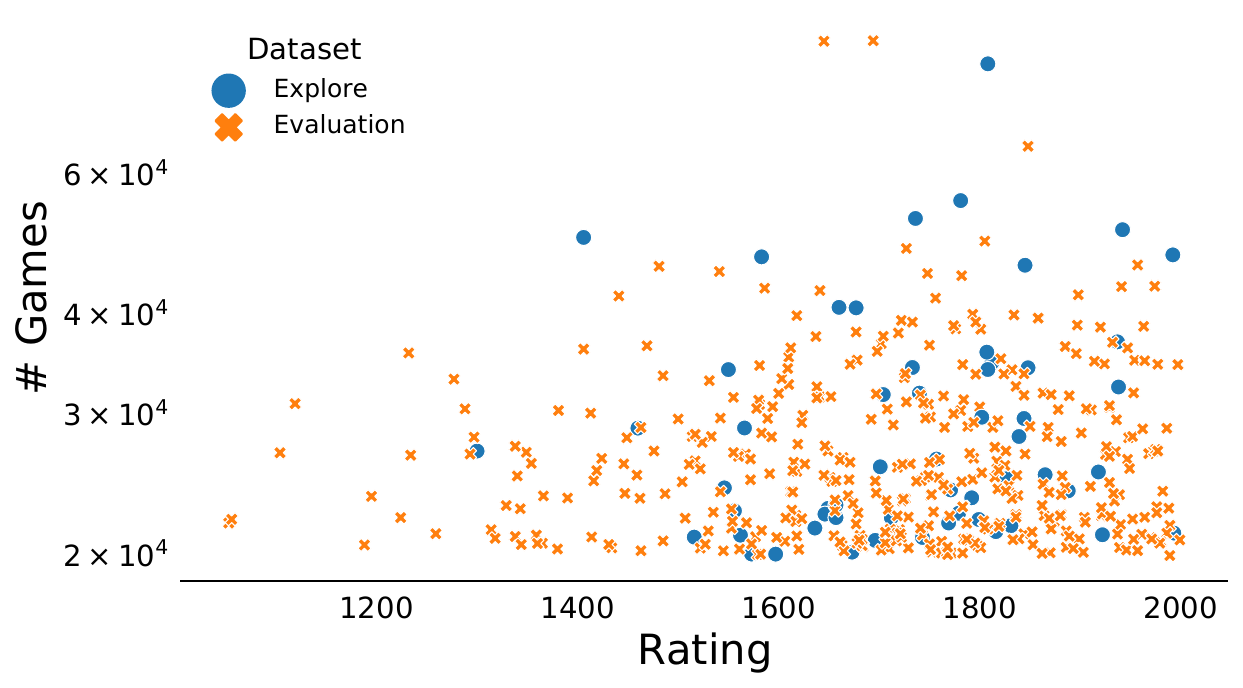}
 \caption{Distribution of players by rating and count }
 \label{data_summary_rating}
\end{figure}

Figure~\ref{data_summary_rating} shows the number of games as a function of rating for each player with 20,000 or more games in the final analysis.

We include the final number of games used for each player's partitions, which are given in Table~\ref{med_count}. 

\begin{table}[h]
    \centering

\begin{tabular}{llrrrrrr}
\toprule
Player & Game &  1000 &  5000 &  10000 &  20000 &  30000 &  40000 \\
\midrule
   Explore &    train &   936 &  3399 &   5308 &   7488 &   9662 &  14086 \\
   Explore &     test &   131 &   454 &    716 &    978 &   1256 &   1796 \\
   Explore &      val &   117 &   417 &    636 &    914 &   1180 &   1710 \\
   Explore &   future &    78 &   264 &    362 &    464 &    956 &    888 \\
  Eval &    train &  1176 &  2432 &   5081 &   7587 &  10292 &  13116 \\
  Eval &     test &   160 &   380 &    701 &    992 &   1336 &   1688 \\
  Eval &      val &   144 &   344 &    630 &    910 &   1252 &   1600 \\
  Eval &   future &   116 &   186 &    347 &    507 &    631 &    808 \\
\bottomrule
\end{tabular}
\caption{Median number of games in each dataset per player partition}\label{med_count}
\end{table}

The median dates are also provided (see Table~\ref{med_date}), which demonstrate the temporal separation between the future set and other 3 sets.

\begin{table}[ht]
\begin{tabular}{llrrrrrr}
\toprule
Player & Game &  1000 &  5000 &  10000 &  20000 &  30000 &  40000 \\
\midrule
 Explore &    train &  20-03 &  20-02 &  20-01 &  19-12 &  19-12 &  19-12 \\
 Explore &     test &  20-03 &  20-02 &  20-01 &  19-12 &  20-01 &  19-12 \\
 Explore &      val &  20-03 &  20-02 &  20-01 &  19-12 &  20-01 &  19-12 \\
 Explore &   future &  20-12 &  20-12 &  20-12 &  20-12 &  20-12 &  20-12 \\
Eval &    train &  20-02 &  20-01 &  20-01 &  19-12 &  19-12 &  19-12 \\
Eval &     test &  20-02 &  20-01 &  20-01 &  20-01 &  19-12 &  19-12 \\
Eval &      val &  20-01 &  20-01 &  20-01 &  19-12 &  19-12 &  19-12 \\
Eval &   future &  20-12 &  20-12 &  20-12 &  20-12 &  20-12 &  20-12 \\
\bottomrule
\bottomrule
\end{tabular}
\caption{Median date from each game in each dataset per player partition}\label{med_date}
\end{table}

\subsection{Methods appendix}

\begin{figure}[t]
 \includegraphics[width=.45\textwidth]{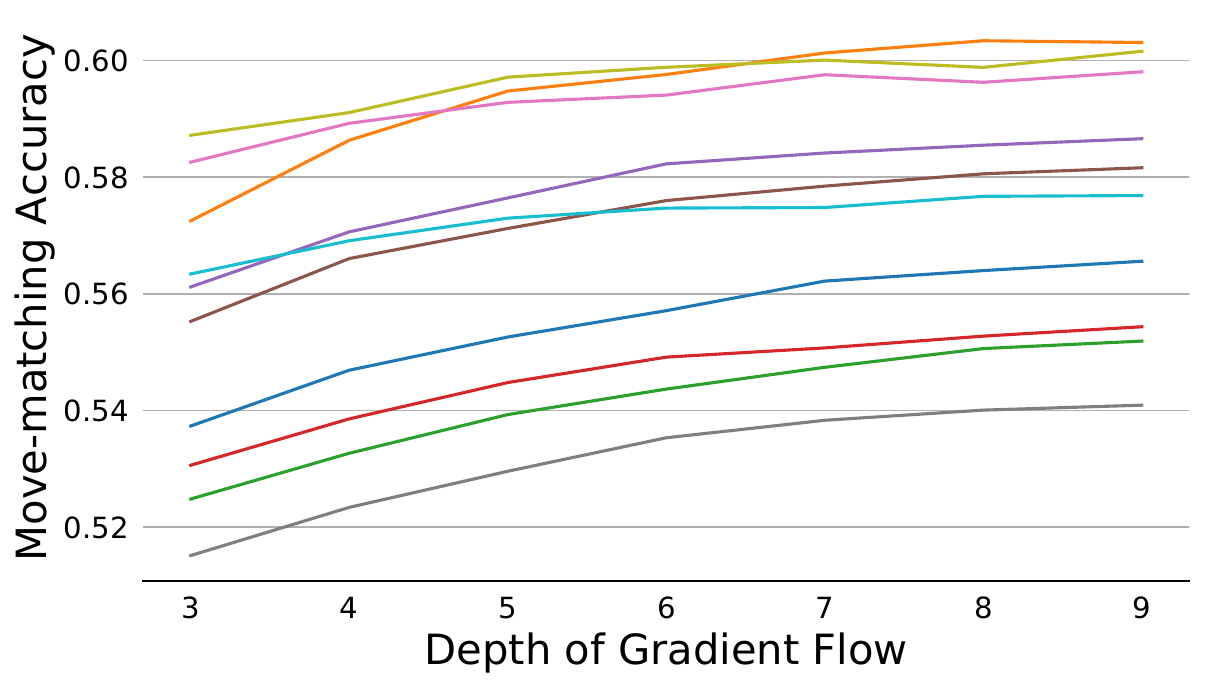}
 \caption{Final accuracy of each player in the 40,000 games explore players set at all possible depths of gradient flow. Each line is a different player.}
 \label{gradient_depth}
\end{figure}

On the final model training procedure Figure \ref{gradient_depth} shows the effects of changing the gradient flow depth on final accuracy.

Table \ref{explore_accuracy} shows the accuracy of on exploration players under different parameters.

\begin{table*}[th]
	\centering
	\begin{tabular}{lrrrrrrrrr}
\toprule
\# Player's Games &  Maia 1100 &  Maia 1500 &  Maia 1900 &  Maia nearest &  LR = .000001 &  LR = .00001 &  LR = .0001 &  LR = .001 &  150,000 Steps \\
\midrule
10,000                &   0.490 &   0.526 &   0.526 &      0.535 &              0.538 &             0.570 &            0.578 &           0.507 &             0.573 \\
20,000                &   0.480 &   0.516 &   0.519 &      0.524 &              0.546 &             0.574 &            0.580 &           0.538 &             0.575 \\
30,000                &   0.475 &   0.517 &   0.529 &      0.530 &              0.553 &             0.581 &            0.590 &           0.555 &             0.582 \\
40,000                &   0.494 &   0.528 &   0.540 &      0.544 &              0.553 &             0.580 &            0.593 &           0.565 &             0.582 \\
\bottomrule
\end{tabular}

	\caption{Move-matching accuracy of exploration player set during final tuning. \textit{LR = .0001} is the final configuration used in Section \ref{sec:results}}
	\label{explore_accuracy}
\end{table*}

As we mention in the main text, there is no significant difference between the future and test set accuracy. Accuracy on the future set is very slightly higher than on the test set, as shown in Table \ref{delta_future}.

\begin{table}[ht]
    \centering
\begin{tabular}{lr}
\toprule
Count &                 Change in Accuracy\\
\midrule
1000                 &         0.0112 \\
5000                 &         0.0114 \\
10000                &         0.0073 \\
20000                &         0.0061 \\
30000                &         0.0050 \\
40000                &         0.0045 \\
\bottomrule
\end{tabular}

    \caption{Increase in accuracy between the testing set and future set for the different player categories.}
    \label{delta_future}
\end{table}

As all our displayed results are based on counts so to get a confidence interval we can though treat each individual player as a different trial. This then gives main effect of the move-matching difference between Maia 1900 and personalized Maia as 4.186\%, with a confidence interval (standard error) of +/- 0.101\%. 

We primarily trained our models on 6 core virtual machines with Tesla V100 GPUs. The training time for one model was 40 minutes when run alone. 
As the models are relatively small (using $\sim 300$MB of vRAM), our training procedure should be feasible on most GPUs.
\subsection{Additional Plots}

\subsubsection{Starting Maia Choice}\label{sec:sub_heatmap}

\begin{figure}[t]
 \centering
 \includegraphics[width=.45\textwidth]{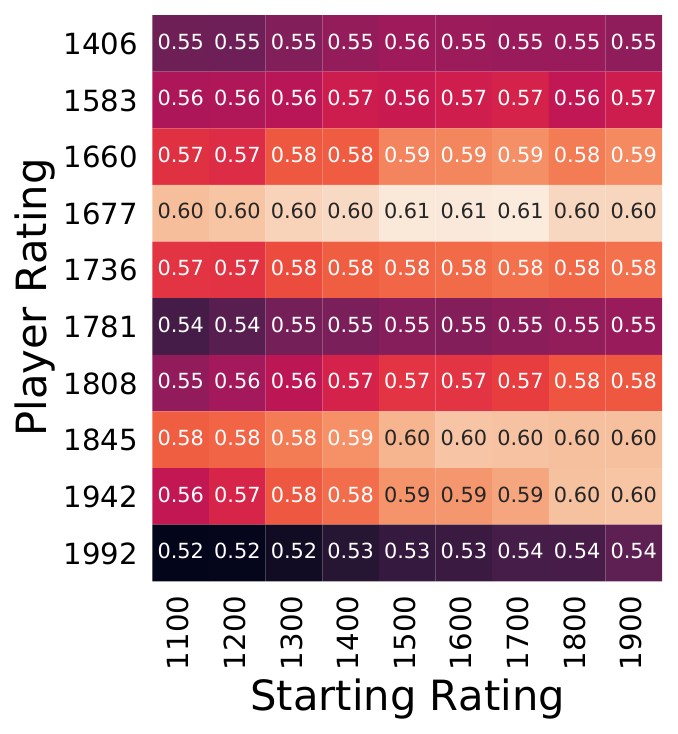}
 \caption{Test accuracy resulting from starting with base Maia level $x$ and fine-tuning to a sample player of rating $y$. }
 \label{rating_heartmap}
\end{figure}

Figure \ref{rating_heartmap} shows the results of our experiments looking at the choice of starting Maia for 10 players, as discussed in Section \ref{sec:params}.

 \begin{figure}[t]
 \includegraphics[width=.45\textwidth]{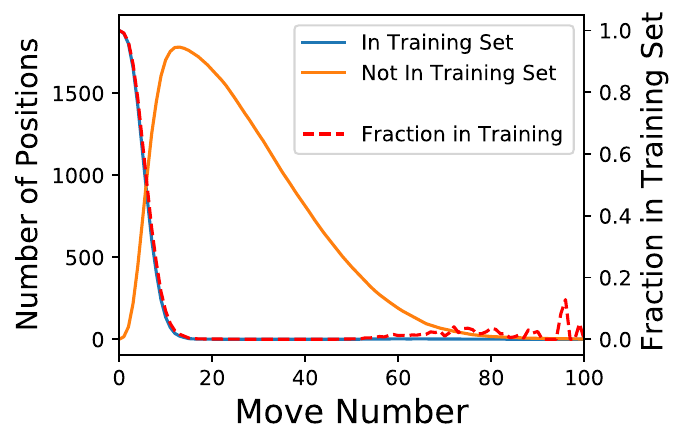}
 \caption{Number of positions in training set by move number. }
 \label{per_ply}
\end{figure}

 \begin{figure}[t]
 \includegraphics[width=.45\textwidth]{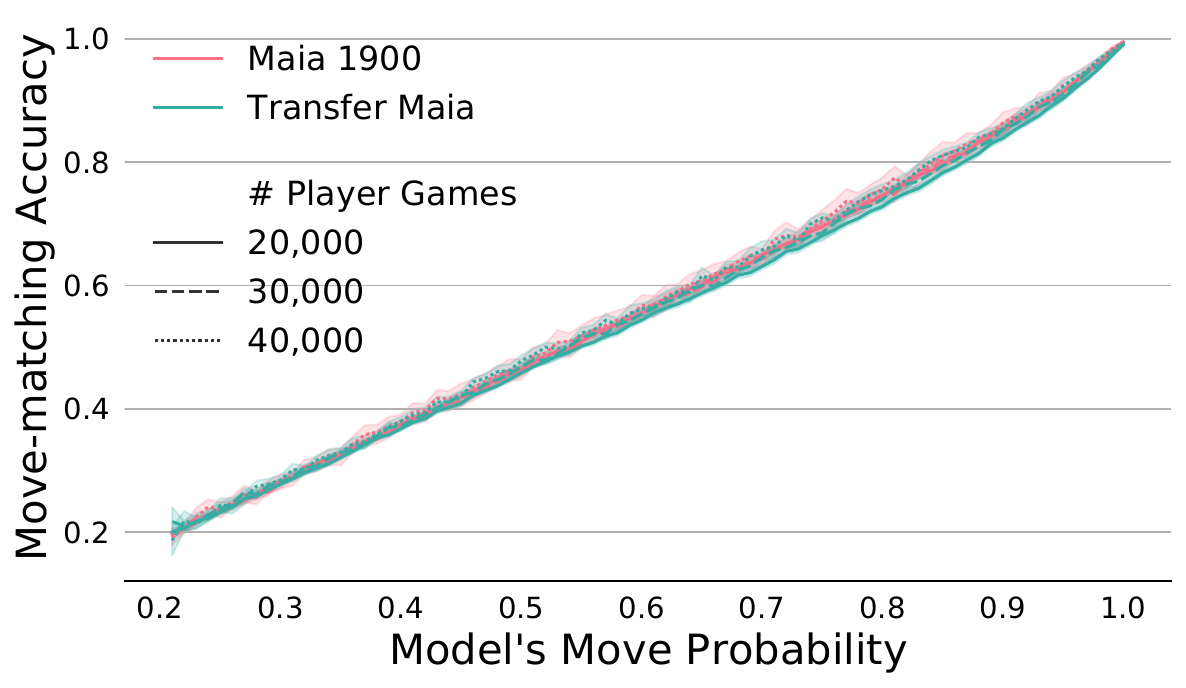}
 \caption{Accuracy vs predicted probability (policy head's raw output) for top move }
 \label{acc_vs_prob}
\end{figure}

 \begin{figure}[t]
\includegraphics[width=.45\textwidth]{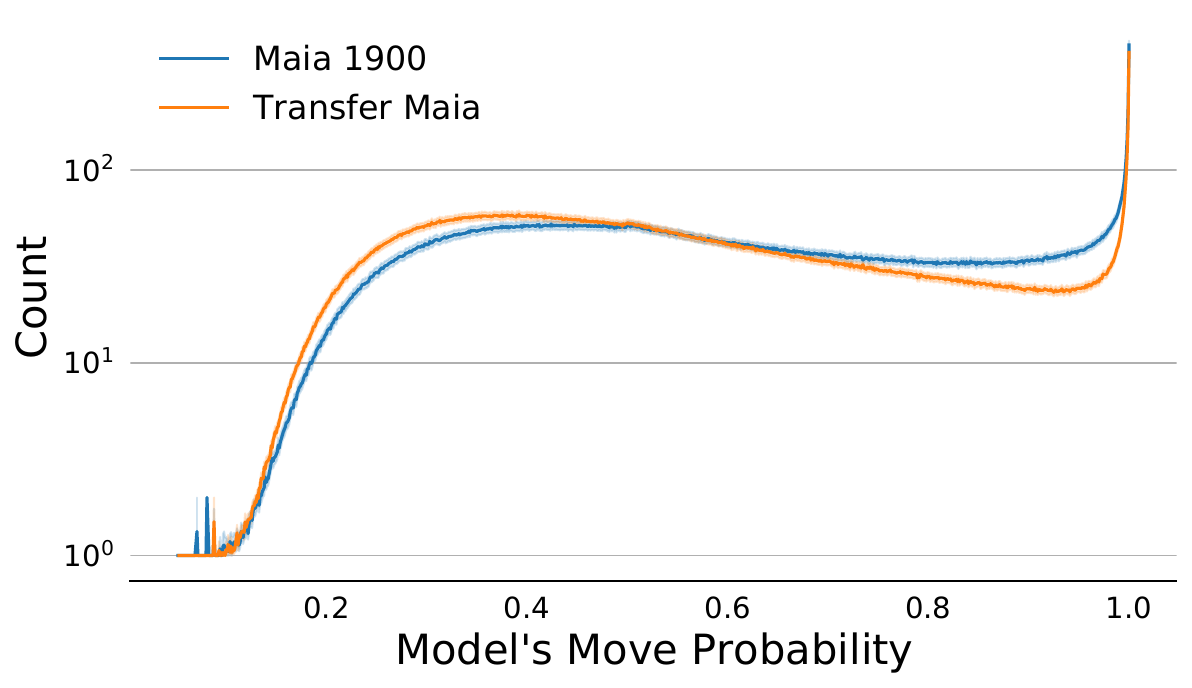}
 \caption{ Distribution of model certainties across all games per player, log Y axis}
 \label{count_vs_policy}
\end{figure}

\subsubsection{Model Calibration}

Our models function by outputting a probability for each move\footnote{The output is an unnormalized distribution over all possible moves that is then normalized with a softmax to only legal moves.}. Instead of only examining the most likely move, we can also consider the top $k$ moves. Figure \ref{acc_vs_prob} shows the accuracy versus predicted probability of the move occurring for both our transfer models and Maia 1900. All models are good at predicting their own accuracy. Figure \ref{count_vs_policy} shows the number of positions each model has a given probability for.

\subsubsection{Stylometry by Ply}

Instead of looking at the stylometry accuracy of our ensemble models on some number of games whole games, we also considered it with when limited to some range of ply within those games.

\subsection{Number of positions per ply per set}

Figure \ref{per_ply} shows the number of positions in the training set as a function of move number ($\frac{ply}{2}$).

\subsection{Stylometry bounds additional details}\label{sec:stylo_supp}

We use two \textit{Chernoff bounds} \cite{chernoff1952measure} for the sum of independent 0-1 random variables. if $X_1, \ldots, X_m$ are independent $0$-$1$ random variables 
whose sum has mean $\mu = \sum_i EX_i$, and if $0 < \delta < 1$, then $
\Prb{X \geq (1 \pm \delta) \mu} < e^{-\delta^2 \mu / 3}$.

Let us use this to bound the probability that $M_i$ (the model trained
on player $A_i$) makes at most $pm (1 + \ve/2)$ correct predictions. The expected number of correct predictions is $mp(1 + \ve)$ (by expanding the expected value), and $pm (1 + \ve) (1 - \ve/2) > pm (1 + \ve/2)$, we can set $\delta = \ve /2$
in bound above and conclude that the probability 
$M_i$ makes at most $pm (1 + \ve/2)$ correct predictions is less than
$e^{-(\ve/2)^2 pm(1 + \ve) / 2} < e^{-\ve^2 pm / 8}$.
Using our assumption that $p \geq 1/2$, this is upper-bounded by
$e^{-\ve^2 m / 16}$.

Next, we bound the probability any other model $M_j$ (trained one of
the other players) makes at least $pm (1 + \ve/2)$ correct predictions.
Since the expected number of correct predictions is $pm$, we can set 
$\delta = \ve/2$ in bound above 
and conclude that the probability
$M_j$ makes at least $pm (1 + \ve/2)$ correct predictions is less than 
$e^{-(\ve/2)^2 pm / 3} < e^{-\ve^2 pm / 12}$.
Again using our assumption that $p \geq 1/2$, this is upper-bounded by
$e^{-\ve^2 m / 24}$.
Finally, taking the union bound over all $n-1$ other players' models,
the probability that any of these models makes at least 
$pm (1 + \ve/2)$ correct predictions is less than 
$(n-1) e^{-\ve^2 m / 24}$.
Now, if none of these bad events happens --- if the correct model doesn't
make at most $pm (1 + \ve/2)$ correct predictions, and if all other models
don't make at least $pm (1 + \ve/2)$ correct predictions --- then 
the model $M_i$ trained on $A_i$ will have the most correct predictions
as desired.
The probability that any of these bad events happens, summing over
the bounds in the previous two paragraphs, is less than equation \ref{eq:bounds}.

\begin{equation}\label{eq:bounds}
(n - 1)  e^{-\ve^2 m / 24} + e^{-\ve^2 m / 16} < n e^{-\ve^2 m / 24}
\end{equation}

We want this to be less than $\delta$ (our accuracy parameter), and
so we need $m$ to satisfy equation \ref{stylo_eq}.
\begin{eqnarray}\label{stylo_eq}
    n e^{-\ve^2 m / 24} & < & \delta \notag\\
    m & > & \frac{24}{\ve^2}\left[\ln n + \ln\left(\frac{1}{\delta}\right)\right]
\end{eqnarray}